\documentclass[11pt,twocolumn]{article}
\usepackage{pslatex, graphicx, amssymb, amsmath}
\usepackage{booktabs}
\usepackage[pagebackref,breaklinks,colorlinks]{hyperref}

\usepackage[capitalize]{cleveref}
\crefname{section}{Sec.}{Secs.}
\Crefname{section}{Section}{Sections}
\Crefname{table}{Table}{Tables}
\crefname{table}{Tab.}{Tabs.}

\def\abstract{
\typeout{Abstract}
 {\bf Abstract} 
}

\begin{document}

\title{Targeted Data Augmentation for bias mitigation}

\author{
Agnieszka Mikołajczyk-Bareła\\
Gdańsk University of Technology\\
{\tt\small agnieszka.mikolajczyk@pg.edu.pl}
\and
Maria Ferlin\\
Gdańsk University of Technology\\\
{\tt\small maria.ferlin@pg.edu.pl}
\and
Michał Grochowski\\
Gdańsk University of Technology\\\
{\tt\small michal.grochowski@pg.edu.pl}
}
\maketitle

\begin{abstract}
The development of fair and ethical AI systems requires careful consideration of bias mitigation, an area often overlooked or ignored. In this study, we introduce a novel and efficient approach for addressing biases called Targeted Data Augmentation (TDA), which leverages classical data augmentation techniques to tackle the pressing issue of bias in data and models. Unlike the laborious task of removing biases, our method proposes to insert biases instead, resulting in improved performance. To identify biases, we annotated two diverse datasets: a dataset of clinical skin lesions and a dataset of male and female faces. These bias annotations are published for the first time in this study, providing a valuable resource for future research. Through Counterfactual Bias Insertion, we discovered that biases associated with the frame, ruler, and glasses had a significant impact on models. By randomly introducing biases during training, we mitigated these biases and achieved a substantial decrease in bias measures, ranging from two-fold to more than 50-fold, while maintaining a negligible increase in the error rate.
\end{abstract}
%
%
%
\section{Introduction}
Data augmentation (DA) finds its application in any deep learning-based system. In computer vision, images are usually augmented by simple, linear transformations, color augmentations, and occasionally more sophisticated techniques, such as cutout or neural transformations \cite{shorten2019survey}. The objective of data augmentation is to improve the efficiency of models analyzed by standard evaluation metrics, such as accuracy, precision, or recall. However, one of the most significant issues in data analysis is the existence of bias, which can distort or undermine results without being immediately apparent \cite{9379034}.

Removing bias from data is often a challenging and time-consuming task, but it is crucial to ensure the validity and fairness of machine learning models. Unfortunately, the process of removing biases, such as eliminating hair from skin lesions or camera reflections from images, can leave behind residual biases or create new artifacts that are difficult to address, even when removed with advanced techniques like image inpainting.

Therefore, we propose the opposite approach: instead of removing biases, we aim to enrich the training sets with selected biases to force the model to learn to ignore them. By randomly adding biases to the input during training, the model will start ignoring them as such features will seem irrelevant. This approach, called Targeted Data Augmentations, breaks the cycle of mistaking correlation with causation by disrupting spurious correlations.

The proposed methodology behind Targeted Data Augmentation (TDA) consists of four steps: bias identification (step 1), augmentation policy design (step 2), training with data augmentation (step 3), and model evaluation (step 4).

In the identification of bias, we utilized a preliminary, supervised step that aimed to detect any possible unwanted biases. To achieve this, we used manual data exploration. We manually labeled 2000 skin lesion images and automatically annotated the entire gender dataset with a trained glasses detection model. Based on the detected biases, we then proposed an augmentation policy that mimicked the biases and injected them into the training data. Additionaly, we provide examples of how to modify features to mitigate bias in different contexts, including tabular data, computer vision, NLP and audio or signal data.

After training the model with this augmented data, we measured the bias using the Counterfactual Bias Insertion (CBI) method \cite{mikolajczyk_towards_2021}. We successfully augmented both the clinical skin lesion dataset and the gender classification dataset by randomly adding black frames and ruler marks to the skin lesion images and inserting eyeglasses into the gender classification images. Our method showed a significant reduction in bias measures, with two to over fifty times fewer images switching classes after the training with TDA, without significantly increasing the error rate.

Our contribution is three-fold: first, we propose a novel bias mitigation method that can easily complement the machine learning pipeline. Second, we present a bias mitigation benchmark that includes two publicly available datasets (over 2000 skin lesion and 50k gender images annotated by us), the code for TDA and bias evaluation, detailed results, and prepared collections of masks and images for bias testing. Third, we identify and confirm previously unknown biases in the gender classification dataset, analyze the robustness of popular models against bias, and show that standard evaluation metrics are often insufficient for detecting systematic errors in data.

The paper is organized as follows: we briefly introduce the topic in the introduction. Next, we describe related works refereed in the paper in the second section, in \cref{sec:tda} we introduce the readers with proposed Targeted Data Augmentation method and analyzed datasets. The details of conducted experiments as well as the results are delivered in \cref{sec:experiments}, and finally, we conclude the research in the last section.

\section{Related works}
 In this paper, we proposed Targeted Data Augmentation method to mitigate selected biases in data for more robust classification. By 'bias in data' we mainly refer to four common data biases in Machine Learning: \textit{observer  bias} \cite{mahtani2018catalogue} which might appear when annotators use personal opinion to label data; \textit{sampling bias} when data is acquired in such a way that not all samples have the same sampling probability \cite{mehrabi2019survey}; \textit{data handling bias} when how data is handled distort the classifier's output; and \textit{instrument bias} meaning imperfections in the instrument or method used to collect the data \cite{he2012bias}.
 
\subsection{Bias mitigation}

It is well documented that models, in most cases reflect the bias in data and often amplify it \cite{zhao2017men}.
Commonly, biases are hardly visible, and their potential impact on the performance of the models is complex and unknown, especially in cases when the model demonstrates high-accuracy results.
The most obvious solution is to remove all biases (e.g features or artifacts) at the pre-processing stage, before training the models. Such an approach called fairness through blindness was proposed by Wang et al. \cite{wang2020towards}.  
It uses the idea that we can remove potentially biasing variable from the input features. For instance, we could remove the information about the candidate's gender when evaluating the potential job candidate's resume. However, in practice, some information about gender might be encoded in the resume, e.g., feminine hobby connected to the gender or gender-specific adjectives. However, removing all potential biases is often very hard, if not impossible, especially in computer vision. One of the applications in which the problem of artifact removal has been analyzed for many years is the issue of analyzing skin lesions for possible cancer detection.  \cite{abbas2011hair,oliveira2016computational}.

Also in other fields and applications, this problem is actively analyzed by many researchers. Zhao et al. \cite{zhao2017men} proposed an inference update scheme to match a target distribution to remove bias. Their method introduces corpus-level constraints so that selected features co-occur no more than in the original training distribution. 
Dwork et al. \cite{dwork2018decoupled} proposed a scheme for decoupling classifiers that can be added to any black-box machine learning algorithm. They can be used to learn different classifiers for different groups. Another branch of approaches is adversarial bias mitigation, i.e., supervised learning
in proposed by Zhang et al.  \cite{zhang2018mitigating} method, the task is to predict an output variable $Y$ given an input variable $X$, while remaining unbiased with respect to some variable $Z$. This approach used the output layer of the predictor as an input to another model called the adversary network that attempts to predict $Z$. The idea was further improved by Le Bras et al. \cite{le2020adversarial}, who proposed the idea of Adversarial Debasing Filters. The proposed algorithm used linear classifiers trained on different random data subsets at each filtering phase. Then, the linear classifier's predictions are collected, predictability score is calculated. High predictability scores are undesirable as their feature representation can be negatively exploited -- hence Le Bras et al. \cite{le2020adversarial} proposed simply removing the top $n$ instances with high scores. The process is then repeated several times to reduce the bias influence.

Finally, there are a number of attempts exploiting the advantages of attention guidance. Early works about attention guidance in computer vision focused on improving the segmentation task \cite{huang2019brain}, making classification better with attention approaches used in Natural Language Processing \cite{barata2019deep}, or even using attention maps to zoom closer to the region of interest \cite{li2019zoom}. The guidance provided with, for example attention maps, highlights relevant regions and suppresses unimportant ones, enabling a better classification. A similar method is based on self-erasing networks that prohibit attention from spreading to unexpected background regions by erasing unwanted areas \cite{hou2018self}. Some researchers proposed different ways to solve this problem, such as rule extraction, built-in knowledge, or built-in-graphs \cite{chai2020human}.


The research presented in this article concerns the bias mitigation by taking the advantage of proposed Targeted Data Augmentation method. To validate the approach we have researched one of the most popular skin lesion benchmarks, namely, ISIC-2020\footnote{
\href{https://www.kaggle.com/nroman/melanoma-external-malignant-256}{Skin Lesion Images for Melanoma Classification}} \cite{isic2020} and gender classification dataset\footnote{ 
\href{https://www.kaggle.com/datasets/cashutosh/gender-classification-dataset}{Gender classification dataset}}. 

\subsection{Datasets}\label{sec:datasets}
There is very little literature focused on defining biases in datasets. Torralba et al. \cite{torralba2011unbiased} examined a cross-dataset generalization on popular benchmarks by evaluating the performance of “car” and “person” classes when training on one dataset and testing on another. 

Regarding the skin lesions classification problem, the existence and influence of bias in the skin lesion datasets have been previously analyzed by the researchers, yet the problem has still not been thoroughly researched, grounded and explained. 
Bissoto et al. \cite{Bissoto_2019_CVPR_Workshops,bissoto_debiasing_2020} conducted research on biases in skin lesion benchmarks and their impact on the quality of model performance. They showed that existing dermoscopy artifacts such as frames, gel bubbles, or ruler marks distort the results and are a common source of bias in data. Van Simoens and Dhoedt \cite{stoyanov_visualizing_2018} proved that the model, in addition to medically relevant features was driven by artifacts such ass pecular reflections, gel application, and rulers. 
In Mikołajczyk et al. \cite{mikolajczyk_biasing_2022} the authors showed that there is a strong correlation between artifacts like black frames, ruler marks, and the skin lesion type (benign/malignant). They showed that models trained on biased data learned spurious correlations resulting in more errors in images with such artifacts.

Finally, in \cite{bissoto_debiasing_2020} Bissoto et al. conducted a comprehensive analysis of 7 visual artifacts and their influence on the deep learning models and employed debiasing methods to decrease their impact on the model's performance. They concluded, that existing state-of-the-art methods for bias removal are unfortunately not capable of handling these biases effectively.


\begin{figure*}[!htb]
\centering
  \includegraphics[width=1.0\textwidth]{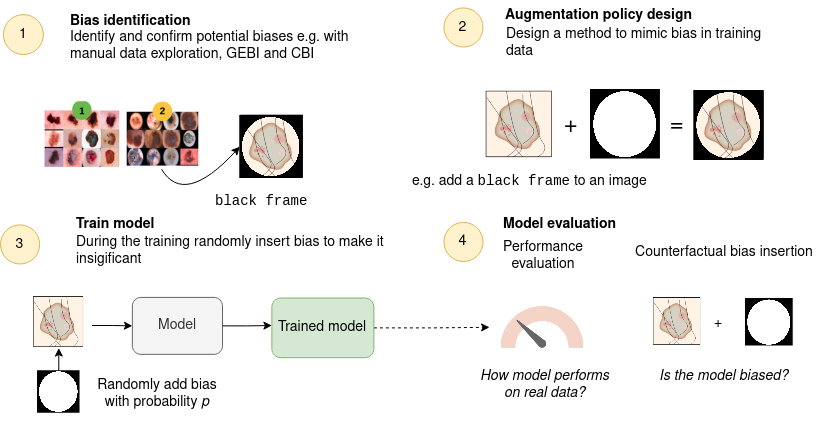}
  \caption{Targeted data augmentation steps.}
  \label{fig.pipeline}
\end{figure*}

\textbf{ISIC-2020} \cite{isic2020, isic2020_metadata} is the largest skin lesion dataset divided into two classes -- benign and malignant. It contains 33,126 dermoscopic images from over 2,000 patients, and the diagnoses were confirmed either by histopathology, expert agreement, or longitudinal follow-up. The dataset was gathered by The International Skin Imaging Collaboration (ISIC) from several medical facilities and was used in the SIIM-ISIC Melanoma Classification Challenge. In the images, the lesion is usually in the center and well-visible. Examples of artifacts that may introduce bias into the model, such as hair, frames, rulers, pen marks, or gel drops, are present in this dataset. Past research has shown that frames are correlated with the malignant class and ruler marks with the benign \cite{mikolajczyk_biasing_2022}.

\textbf{Gender classification dataset} consists of cropped images of male and female faces. The data were collected from various internet sources, most of which were extracted from the IMDB dataset. It contains 58,658 images, with a similar number of images in the female and male subsets. We discovered that glasses could be a potential source of bias, as actors wore them more often than actresses.

\section{Targeted Data Augmentation}
\label{sec:tda}
Within the framework of the proposed {Targeted Data Augmentation} method, the following stages can be specified: bias identification, augmentation policy design, training with targeted data augmentation, and  finally model evaluation. The TDA pipeline is presented in \cref{fig.pipeline}.

The bias identification is a preliminary, supervised step in which we try to detect unwanted bias within the data. Here, we manually explored data to detect bias and then, measured it by using counterfactual bias insertion \cite{mikolajczyk_towards_2021}. The identified biases can be identified, explained and also mitigated using data augmentation methods.  

In the next step, we have to train the model using targeted data augmentation. This process involves randomly adding a given bias to the training data,  according to a designed augmentation policy, e.g. adding a black frame to the skin lesion images or glasses to face images. This process is intended to make biases less correlated with a given class and increase its randomness. Finally, we compared the performance and counterfactual measures of bias insertion process for the trained model with different bias augmentation probabilities.  

\subsection{Bias identification based on the errors of an existing model}\label{sec.bias_identification}

The key to successful bias-targeted data augmentation is thoroughly identifying potential biases and their sources. This can be accomplished through manual inspection of data, using global explanation methods (e.g., Global Explanations for Bias Identification \cite{mikolajczyk_towards_2021}), or by conducting a literature review or consulting with domain experts.

In our research reported in this paper, potential biases were selected through manual data analysis. 
To make the manual data inspection process more objective, we proposed to base it on three following basic metrics: \textit{an artifacts cardinality}, an \textit{artifact ratio} and \textit{a class ratio.} 

\textit{Cardinality of artifacts} refers to the total number of elements (images) where a certain artifact is present within a class. The cardinality cannot exceed the number of annotated images per class.

\textit{Artifact ratio}. The artifact ratio $Q^{artifact}$ is calculated as the number of images with certain artifacts divided by the total number of images investigated. The artifact ratio indicates the percentage of samples with a certain artifact out of all samples. We calculated artifact ratios separately for each class.

\textit{Class ratio.} Class ratio $Q^{class}$ is equal to the ratio of artifact ratios from two classes $C_1$ and $C_2$: 

\begin{equation}
Q^{class, artifact} = \frac{Q^{artifact, C_1}}{Q^{artifact,C_2}}
\end{equation}

A class ratio close to one indicates that both classes have a similar incidence of artifacts. Significantly higher or lower class ratios suggest that the examined artifact is more common in one class than the other.

\subsection{Augmentation policy design}
An augmentation policy should clearly specify \textit{which} feature is being modified and \textit{how} it should be modified. This policy can involve modifying existing features, adding new elements, removing elements, and swapping categorical feature values, and it can be applied to computer vision, tabular data, text, audio, or even signals. For instance, if a potential bias is identified based on a feature like \textit{country of origin}, we can modify the value of the \textit{country of origin} during training by randomly switching \textit{Poland} to \textit{China}. Similarly, when classifying pigeon songs in \textit{bird song classification}, we can randomly add city noises to the samples to mitigate the bias that arises from city sounds since pigeons are often recorded in cities, unlike other bird species. If the algorithm is biased toward a certain \textit{age} range, such as performing well on people aged 16--18 but poorly on 19-year-olds, we can modify \textit{age values} within a certain range to mitigate the bias.

This method can also be applied to natural language processing (NLP) models. For example, if the model is biased towards a certain gender, like classifying "he" more accurately than "she", we can randomly switch gender pronouns. Similarly, if we identify a bias towards a certain race or ethnicity, we can modify the names or other identifying characteristics in the training data.

In addition to these examples, there are other ways to use augmentation to address bias in data. For instance, we can use GANs to modify certain image features such as skin color, change the background of an image from summer to winter, or manipulate the lighting conditions to simulate different environments. By carefully selecting and designing augmentation strategies, we can help our models become more robust to bias and improve their generalization performance on diverse datasets.
 
In this paper, we examine two image classification problems: skin lesion classification and gender classification. To mitigate the potential biases in these datasets, we have implemented augmentation techniques that involve adding ruler marks and black frames to the skin lesion images and adding glasses to the gender images. 

\textbf{Frame augmentation.} Frames, also known as dark corners, are both black and white round markings around the skin lesions, black rectangle edges, and vignettes. We focused on the black round and rectangular markings of different sizes and shapes. Frame augmentation was done by randomly inserting different types of black frames during training. Additionally, each frame was randomly scaled and rotated. We used six different types of frames in each training and a separate set of five frames for the evaluation procedure.

\textbf{Ruler augmentation.} Ruler marks are partially or fully-visible ruler markings of different shapes and colors that can be found throughout the dermoscopic skin lesion datasets. We used pairs of images and segmentation masks of rulers from a designated subset of data to copy rulers from the source image to the target image. This allowed for achieving good augmentation quality without significant loss in computation performance. As in the previous case, we randomly scaled and rotated segmentation masks.

\textbf{Glasses augmentation.} Glasses are objects that may be visible in the face image but do not belong to the face itself. We used masks of different types of glasses, including sunglasses, and randomly inserted them into the image at eye level. In accordance with \cite{twarz_atlas}, we placed them in one-third of a human face. In this case, we did not randomly rotate and scale the mask as it could result in a strange position of glasses in relation to the face. We provided 30 different masks for training and eight other masks for evaluation.

\begin{figure*}[!htb]
\centering
  \includegraphics[width=0.7\textwidth]{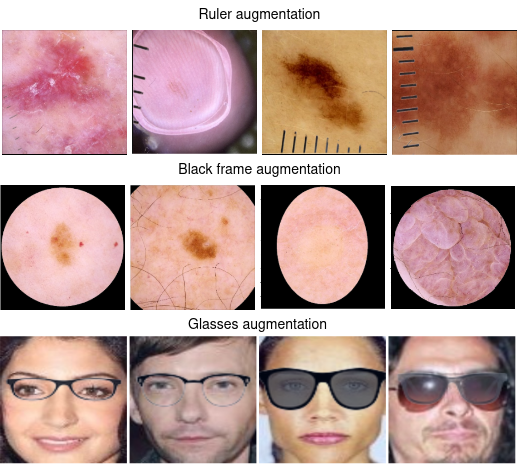}%
  \caption{Example bias augmentations.}
  \label{fig.augmentation-examples}
\end{figure*}

Examples of black frames, rulers, and glasses augmentation are presented in \cref{fig.augmentation-examples}.

\subsection{Training with Targeted Data Augmentation}
Training with targeted data augmentation (TDA) is similar to classical training, but requires a specific augmentation method designed to address bias. TDA can be combined with other data augmentation and bias mitigation techniques. To mitigate spurious correlations between selected features and outputs, TDA should be applied randomly with a designated probability $p$.

\subsection{Bias Evaluation} \label{sec.evaluation}

Bias evaluation is a crucial step in bias mitigation pipelines. We utilized the Counterfactual Bias Insertion (CBI) method \cite{mikolajczyk_towards_2021} to measure the influence of bias. The CBI method involves comparing the model's prediction on the original input with its prediction on the same input with the examined bias inserted. To accomplish this, we first computed the predictions for all samples in the dataset and stored them. Then, we inserted the bias into every sample, computed the predictions for all biased samples, and compared them with the original predictions.

\begin{figure}[!htb]
\centering 


 \includegraphics[width=0.4\textwidth]{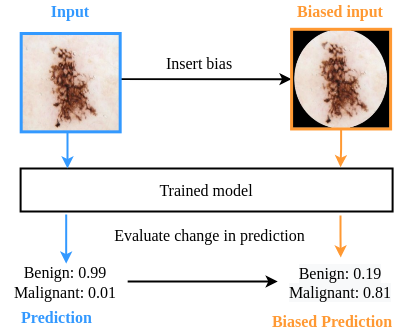}%

 \caption{Counterfactual bias insertion example.} \label{Figure.counter-bias-image}
 
\end{figure}

Ideally, the model's prediction should remain unchanged after inserting minor artifacts or making small data shifts. An example of inserting black frames is presented in \cref{Figure.counter-bias-image}.

The basic measure is to check the difference between the values of $F_1$ on the original samples and compare it with the values of $F_1^{aug}$ calculated for samples with inserted bias.
Ideally, $F_1^{aug}$ should be the same as $F_1$, what means that adding the bias to the data does not change its performance. Significant differences between $F_1^{aug}$ and $F_1$ mean more susceptibility to bias. The $F_1^{mean}$ index is the average of $F_1$ and $F_1^{aug}$, hence it shows how well the model performs on both original and modified data. 

An additional measure is the number of switched classes. A prediction is considered $switched$ when the predicted class has changed.  If the dataset is not biased with the artifact under study, the number of switched classes should remain zero. The number of reversed classes does not test the accuracy or correctness of the predicted category, but only discusses the effect of bias on the prediction.

Switched class is defined as follows (\cref{eq.switched}):
\begin{equation} \label{eq.switched}
{
    switched(k) = \begin{cases}
        1,& \text{if } c^{out}_{p_k} \neq c^{biased}_{p_k}\\
        0,              & \text{otherwise}
    \end{cases}
}
\end{equation}
where $c^{out}_{p_k}$ is a $class$ (based on prediction $p_k$) for input $k$, $c^{biased}_{p_k}$ is predicted $class$ for biased input $k$, and $n$ is number of all examined predictions.

\section{Experiments}
\label{sec:experiments}
Experiments were conducted to measure the extent to which Targeted Data Augmentation (TDA) helps mitigate biases, how the augmentation probability affects the results, and whether the results differ between the considered models. Following the TDA pipeline, we first carefully identified potential biases: frames and rulers for the skin lesion dataset, and glasses for gender classification. After bias identification, we measured the influence of the detected biases using the Counterfactual Bias Insertion method \cite{mikolajczyk_towards_2021}, which involves adding artifacts to all images and measuring how the predictions change. In the case of an ideal classifier, the addition of artifacts such as a black frame or a ruler mark in the image of a skin lesion, or eyeglasses in the image of a person should not change the prediction. As the influence of the bias was confirmed, we began the bias mitigation process by conducting training with targeted data augmentations.

After completing the bias identification step, we proceeded with the augmentation policy design. We used the augmentations described in the Bias Identification section \cref{sec.bias_identification}. For each dataset, we prepared a separate subset of biases for training and testing to avoid data leakage. For training, we allowed for rotations, zoom in/out, and other modifications of the masks. Each bias was inserted with a certain probability $p$, depending on the training. As we tested two different biases on the skin lesion dataset, we augmented it separately.

We trained three models of different sizes on the same dataset with identical hyperparameters: once classical and once with bias augmentation. During evaluation, we compared the results on clean images without the bias ($p=0.0$) and with inserted biases ($p=1.0$) without modifying the masks to avoid unnecessary randomness. The reported results were averaged for each bias. We used the metrics introduced in the evaluation section \cref{sec.evaluation}, including the $switched$ metric, which measures how many predictions changed after adding the bias to the data.

\subsection{Skin lesion classification}
In terms of skin lesion classification, previous research has shown that black frames are often associated with malignant lesions, whereas ruler marks tend to be correlated with the benign class \cite{mikolajczyk_biasing_2022}. Our own research has further supported these findings, as we have manually labeled an additional 2000 samples, and all of the artifact annotations will be made publicly available. We have combined the results of our annotations with the public labels from \cite{mikolajczyk_biasing_2022}, and the aggregated results can be found in \cref{tab:artifacts.annotation_stats}. 

\begin{table}[!htb]
\centering
\label{tab:artifacts.annotation_stats}
\resizebox{\linewidth}{!}{%
\begin{tabular}{llllll} 
\toprule
type & $\lvert ben \rvert$  &    $Q^{artifact}$     & $\lvert mal \rvert$ &    $Q^{artifact}$     & $Q^{class}$  \\
\midrule
\textbf{frame}    & 104    & 5.20\%  & 521       & 26.05\% & 5.01             \\
hair     & 958    & 47.88\% & 868       & 43.40\% & 0.91             \\
\textbf{ruler}    & 422    & 21.09\% & 586       & 29.30\% & 1.39             \\
others    & 426    & 21.29\% & 818       & 40.90\% & 1.92             \\
none     & 538    & 26.89\% & 268       & 13.40\% & 0.50             \\ 
\midrule
total    & 2001   &         & 2000      &         &                  \\
\bottomrule
\end{tabular}
}
\caption{Aggregated manual annotations from \cite{mikolajczyk_biasing_2022} and manually annotated artifacts in the skin lesion dataset ISIC 2019 \cite{isic2019BCN20000,isic2019codella,isic2019ham10000}, and ISIC 2020 \cite{isic2020}.}
\end{table}

We compared three different measures of bias in \cref{sec.bias_identification}: the \textit{cardinality of artifacts}, the \textit{artifact ratio}, and the \textit{class ratio}. Our analysis revealed that frames are not only a common artifact, but are also strongly correlated with malignancy class. Specifically, frames were present in approximately 26\% of malignant cases and only 5\% of benign cases, resulting in a malignancy class ratio ($Q^{class}$) of approximately 5.01. Ruler marks, while more common than frames, were less correlated with malignancy class, with a $Q^{class}$ of approximately 1.39.

To investigate how bias affects machine learning models, we conducted CBI experiments. We inserted five different types of frames or ruler marks into each test image and measured the impact on model performance. The results, presented in \cref{tab:results-lesions}, showed that all three architectures were strongly affected by the insertion of frames. The difference in $F_1$ score after bias insertion was lowest for EfficientNet, with a score of $F_1^{diff} = 12.7\%$, and highest for ViT, with a score of $F_1^{diff} = 56.62\%$. Predictions were strongly influenced by the inserted biases, with almost all samples switching from benign to malignant class after the insertion of a frame. Ruler insertion also resulted in a drop in accuracy, although not as severe as with frames.

To mitigate the bias observed in our experiments, we used segmentation masks with ruler marks and six masks of frames during training. Training details are presented in \cref{sec.training_details}, and the results are presented in \cref{tab:results-lesions}. Our analysis showed that targeted data augmentation resulted in a significant decrease in switched predictions and an increase in $F_1^{aug}$, and sometimes even in $F_1$. In particular, the effect of frame bias in DenseNet was reduced by a factor of 38, from almost 2000 cases to only 50. For EfficientNet, the effect of frame bias was reduced by a factor of 2.7, and for ViT, it was reduced by a factor of 57. Similarly, the bias introduced by ruler marks was mitigated, with a significant decrease in switched values observed for all three architectures.

\begin{table*}
\centering
\label{tab:results-lesions}
\resizebox{\textwidth}{!}{%
\begin{tabular}{c|ccrrrrrrr} \toprule
\multicolumn{1}{c}{\textbf{model}} & \textbf{type}  & \textbf{p} & \multicolumn{1}{c}{\textbf{switched}} & \multicolumn{1}{c}{\textbf{}} & \multicolumn{1}{c}{\textbf{mal to ben}} & \multicolumn{1}{c}{\textbf{ben to mal}} & \multicolumn{1}{c}{$F_1$} & \multicolumn{1}{c}{\textbf{$F_1^{aug}$}} & \multicolumn{1}{l}{\textbf{$F_1^{diff}$}}  \\ \midrule
\textbf{DenseNet121}               & \textbf{frame} & 0          & 1928.8                                & 25.61\%                       & 0.79\%                                  & 99.21\%                                 & 88.18\%                 & 51.88\%                                  & 36.30\%                                    \\
\textbf{}                          & \textbf{}      & 0.25       & 68.0                                  & 0.90\%                        & 66.76\%                                 & 33.24\%                                 & 87.79\%                 & 85.50\%                                  & 2.29\%                                     \\
\textbf{}                          & \textbf{}      & 0.5        & \textbf{49.8}                         & \textbf{0.66\%}               & 59.84\%                                 & 40.16\%                                 & 88.22\%                 & 87.11\%                                  & 1.11\%                                     \\
\textbf{}                          & \textbf{}      & 0.75       & 82.0                                  & 1.09\%                        & 11.95\%                                 & 88.05\%                                 & \textbf{88.54\%}        & \textbf{88.29\%}                         & \textbf{0.24\%}                            \\
\textbf{}                          & \textbf{}      & 1          & 73.6                                  & 0.98\%                        & 70.65\%                                 & 29.35\%                                 & 87.51\%                 & 86.66\%                                  & 0.85\%                                     \\ \cmidrule(lr){2-10}
\textbf{}                          & \textbf{ruler} & 0          & 142.4                                 & 1.89\%                        & 78.51\%                                 & 21.49\%                                 & 89.62\%                 & 84.82\%                                  & 4.80\%                                     \\
\textbf{}                          & \textbf{}      & 0.25       & \textbf{77.2}                                  & \textbf{1.03\%}                        & 86.01\%                                 & 13.99\%                                 & 89.84\%                 & 88.31\%                                  & 1.52\%                                     \\
\textbf{}                          & \textbf{}      & 0.5        & 92                                    & 1.22\%                        & 90.87\%                                 & 9.13\%                                  & 88.92\%                 & 88.58\%                                  & 0.34\%                                     \\
\textbf{}                          & \textbf{}      & 0.75       & 77.8                                  & 1.03\%                        & 83.03\%                                 & 16.97\%                                 & \textbf{90.40\% }                & \textbf{89.01\%}                                  & 1.39\%                                     \\
\textbf{}                          & \textbf{}      & 1          & 86.6                                  & 1.15\%                        & 90.30\%                                 & 9.70\%                                  & 88.60\%                 & 88.60\%                                  & \textbf{-0.01\%}                                    \\ \midrule
\textbf{EfficientNet-B2}          & \textbf{frame} & 0          & 504.2                                 & 6.70\%                        & 4.72\%                                  & 95.28\%                                 & \textbf{86.56\%}        & 73.81\%                                  & 12.76\%                                    \\
\textbf{}                          & \textbf{}      & 0.25       & \textbf{187.0}                        & \textbf{2.48\%}               & 8.34\%                                  & 91.66\%                                 & 83.23\%                 & \textbf{78.61\%}                         & 4.62\%                                     \\
\textbf{}                          & \textbf{}      & 0.5        & 310.0                                 & 4.12\%                        & 12.97\%                                 & 87.03\%                                 & 79.06\%                 & 73.22\%                                  & 5.84\%                                     \\
\textbf{}                          & \textbf{}      & 0.75       & 411.8                                 & 5.47\%                        & 41.04\%                                 & 58.96\%                                 & 78.12\%                 & 78.00\%                                  & \textbf{0.11\%}                            \\
\textbf{}                          & \textbf{}      & 1          & 204.8                                 & 2.72\%                        & 9.57\%                                  & 90.43\%                                 & 73.02\%                 & 72.90\%                                  & 0.12\%                                     \\ \cmidrule(lr){2-10}
\textbf{}                          & \textbf{ruler} & 0          & 173.4                                 & 2.30\%                        & 97.00\%                                 & 3.00\%                                  & 87.89\%                 & 76.67\%                                  & 11.22\%                                    \\
\textbf{}                          & \textbf{}      & 0.25       & 167.2                                 & 2.22\%                        & 95.81\%                                 & 4.19\%                                  & 87.92\%                 & 79.39\%                                  & 8.52\%                                     \\
\textbf{}                          & \textbf{}      & 0.5        & 157.4                                 & 2.09\%                        & 92.25\%                                 & 7.75\%                                  & 88.97\%                 & 81.64\%                                  & 7.33\%                                     \\
\textbf{}                          & \textbf{}      & 0.75       & 171.2                                 & 2.27\%                        & 93.57\%                                 & 6.43\%                                  & 85.48\%                 & 75.67\%                                  & 9.81\%                                     \\
\textbf{}                          & \textbf{}      & 1          & \textbf{93}                           & \textbf{1.24\%}               & 86.67\%                                 & 13.33\%                                 & 88.56\%                 & \textbf{86.55\%}                         & \textbf{2.01\%}                            \\ \midrule
\textbf{ViT}   & \textbf{frame} & 0          & 5014.2                                & 66.59\%                       & 0.00\%                                  & 100.00\%                                & 88.85\%                 & 32.22\%                                  & 56.62\%                                    \\
\textbf{}                          & \textbf{}      & 0.25       & 187.0                                 & 2.48\%                        & 0.08\%                                  & 91.66\%                                 & 83.23\%                 & 78.61\%                                  & 4.62\%                                     \\
\textbf{}                          & \textbf{}      & 0.5        & 161.2                                 & 2.14\%                        & 2.61\%                                  & 97.39\%                                 & 90.18\%                 & 86.88\%                                  & 3.30\%                                     \\
\textbf{}                          & \textbf{}      & 0.75       & 115.8                                 & 1.54\%                        & 63.90\%                                 & 36.10\%                                 & 88.33\%                 & 87.77\%                                  & \textbf{0.55\%}                            \\
\textbf{}                          & \textbf{}      & 1          & \textbf{87.2}                         & \textbf{1.16\%}               & 60.55\%                                 & 39.45\%                                 & \textbf{90.30\%}        & \textbf{88.49\%}                         & 1.81\%                                     \\ \cmidrule(lr){2-10}
\textbf{}                          & \textbf{ruler} & 0          & 189.4                                 & 2.52\%                        & 93.24\%                                 & 6.76\%                                  & 88.85\%                 & \textbf{87.47\%}                         & 1.38\%                                     \\
\textbf{}                          & \textbf{}      & 0.25       & 37.4                                  & 0.50\%                        & 63.64\%                                 & 36.36\%                                 & 79.87\%                 & 78.99\%                                  & 0.88\%                                     \\
\textbf{}                          & \textbf{}      & 0.5        & \textbf{37}                           & \textbf{0.49\%}               & 69.19\%                                 & 30.81\%                                 & 80.30\%                 & 80.22\%                                  & \textbf{0.08\%}                            \\
\textbf{}                          & \textbf{}      & 0.75       & 290                                   & 3.85\%                        & 96.97\%                                 & 3.03\%                                  & 78.94\%                 & 76.82\%                                  & 2.12\%                                     \\
\textbf{}                          & \textbf{}      & 1          & 33.6                                  & 0.45\%                        & 25.00\%                                 & 75.00\%                                 & 78.76\%                 & 78.54\%                                  & 0.22\%                                     \\ \bottomrule
\end{tabular}
}
\caption{Counterfactual bias insertion results on frame and ruler bias testing with and without Targeted Data Augmentation.  "p": the probability of augmentation, "switched": the number of images where the model switched its prediction when the glasses were added, "mal to ben": the percentage of images switched from benign to malignant, "ben to mal": the percentage of images switched from malignant to benign ,, "$F_1$": the overall $F_1$ score of the model on the task, "$F_1^{aug}$": the $F_1$ score of the model on augmented (biased) data. "$F_1^{diff}$": the difference between the $F_1$ score and the $F_1^{aug}$ score. This is a measure of how much the model benefits from using augmented data.}
\end{table*}

\subsection{Gender classification}
 
To assess the extent of bias towards glasses in the gender classification dataset, we annotated a small subsample of the dataset for the eyeglasses presence classification task. This annotated dataset was used along with the GAN generated dataset \footnote{\textit{\href{https://www.kaggle.com/datasets/jeffheaton/glasses-or-no-glasses}{Glasses or No Glasses}}: a Kaggle dataset from the course \textit{T81-855: Applications of Deep Learning at Washington University in St. Louis }} to train the EfficientNet-B2 model to identify images with glasses. The final glasses presence classifier achieved a high performance score of approximately $F_1 \approx 0.96$. We then automatically annotated the gender dataset\footnote{The glasses annotations will be publicly available} and compared the metrics between samples in both gender categories. This confirmed our suspicion of bias in the dataset, and the results are presented in \cref{tab:artifacts.glasses}.

Although glasses are not very common in this dataset, with less than 12\% of images including glasses for men and slightly over 1\% for women, they are more than seven times ($Q^{class} \approx{7.79}$) more common in images with men than with women. This makes it the strongest single trait disparity between classes in our comparison.

\begin{table}[!htb]
\centering
\label{tab:artifacts.glasses}
\resizebox{\linewidth}{!}{%
\begin{tabular}{llllll} \toprule
type       & $\lvert mal \rvert$  &   $Q^{artifact}$       & $\lvert fem \rvert$&    $Q^{artifact}$     &   $Q^{class}$    \\ \midrule
glasses    & 2659  & 11.19\%  & 334    & 1.44\%  & 7.79  \\
none & 21107 & 88.81\% & 22909  & 98.56\% & 0.90  \\  \hline
total      & 23766 &         & 23243  &         &       \\ \bottomrule
\end{tabular}
}
\caption{Artifacts statistics for semi-automatically annotated gender classification dataset}
\end{table}

As the total number of glasses in the dataset is relatively small, we conducted additional CBI experiments to verify our suspicions. For each test image, we inserted nine different types of glasses and measured the resulting bias. The averaged results (probability $p=0.0$) are presented in \cref{tab:results-gender}.

We confirmed that, similar to the skin lesion dataset, glasses insertion affected all models, despite it being a feature that should not affect the results. After the bias was inserted, each model had a lower $F_1$ score: $F_1^{diff}$ was equal to $4.93$ for DenseNet, $4.6$ for EfficientNet, and $1.21$ for ViT. Consequently, almost all samples that changed predictions switched from the female to male class, once again confirming the correlation between glasses and gender.

To test the bias mitigation algorithm on this dataset, we used 30 masks with both corrective glasses and sunglasses.
The training details are presented in \cref{sec.training_details}, and the results of the training are shown in \cref{tab:results-gender}.
In all cases tested, training with Targeted Data Augmentation resulted in a significant decrease in switched predictions, as well as an increase in $F_1^{aug}$, and sometimes even in $F_1$.
The frame bias in DenseNet, as represented by the $switched$ metric, decreased 3.8 times, 3.3 times for EfficientNet, and 1.8 times for ViT.
These results demonstrate that TDA can have a positive effect, even in cases where bias does not occur very often.

\begin{table*}[!htb]
\centering
\label{tab:results-gender}
\resizebox{\textwidth}{!}{%
\begin{tabular}{@{}cccrrrrrrr@{}}
\toprule
\textbf{model}                                        & \textbf{type} & \textbf{p} & \multicolumn{1}{c}{\textbf{switched}} & \multicolumn{1}{c}{\textbf{}} & \multicolumn{1}{c}{\textbf{mal to fem}} & \multicolumn{1}{c}{\textbf{fem to mal}} & \multicolumn{1}{c}{\textbf{$F_1$}} & \multicolumn{1}{c}{\textbf{$F_1^{aug}$}} & \multicolumn{1}{l}{\textbf{$F_1^{diff}$}} \\ \midrule
\multicolumn{1}{c|}{\textbf{DenseNet121}}             & \textbf{glasses}     & 0          & 908.4                               & 7.8\%                       & 9.98\%                                  & 90.02\%                                 & 96.90\%                         & 91.98\%                              & 4.93\%                               \\
\multicolumn{1}{c|}{\textbf{}}                        & \textbf{}          & 0.25       & \textbf{235.6}                                  & \textbf{2.02\%}                        & 55.99\%                                 & 44.01\%                                 & 96.95\%                         & \textbf{95.88\%}                              & \textbf{1.07\%}                                \\
\multicolumn{1}{c|}{\textbf{}}                        & \textbf{}          & 0.5        & 284.8                         & 2.44\%               & 44.56\%                                 & 55.44\%                                 & \textbf{98.78\%}                         & 95.59\%                              & 1.18\%                                \\
\multicolumn{1}{c|}{\textbf{}}                        & \textbf{}          & 0.75       & 282.9                                  & 2.43\%                        & 53.02\%                                 & 46.98\%                                 & 96.19\%                & 95.27\%                     & 0.92\%                       \\
\multicolumn{1}{c|}{\textbf{}}                        & \textbf{}          & 1          & 337.4                                  & 2.90\%                        & 26.24\%                                 & 73.76\%                                 & 93.42\%                         & 94.14\%                              & 0.71\%                                \\ 
\midrule
\multicolumn{1}{c|}{\textbf{EfficientNet-B2}}        & \textbf{glasses}     & 0          & 800.9                                 & 6.88\%                        & 21.24\%                                  & 78.76\%                                 & 96.41\%                & 91.81\%                              & 4.60\%                               \\
\multicolumn{1}{c|}{\textbf{}}                        & \textbf{}          & 0.25       & 300.7                       & 2.58\%               & 54.36\%                                  & 45.64\%                                 & \textbf{96.90\%}                         & \textbf{95.76\%}                     & 1.15\%                                \\
\multicolumn{1}{c|}{\textbf{}}                        & \textbf{}          & 0.5        & 273.6                                 & 2.35\%                        & 53.57\%                                 & 46.43\%                                 & 96.62\%                         & 95.49\%                              & 1.12\%                                \\
\multicolumn{1}{c|}{\textbf{}}                        & \textbf{}          & 0.75       & \textbf{237.4}                                 & \textbf{2.04\%}                        & 39.07\%                                 & 60.93\%                                 & 96.66\%                         & 95.69\%                              & 0.97\%                       \\
\multicolumn{1}{c|}{\textbf{}}                        & \textbf{}          & 1          & 333.7                                 & 2.86\%                        & 63.97\%                                  & 36.03\%                                 & 95.59\%                         & 95.45\%                              & \textbf{0.13\%}                                                                \\ \midrule
\multicolumn{1}{c|}{\textbf{ViT}} & \textbf{glasses}     & 0          & 262.1                                & 2.25\%                       & 11.45\%                                  & 88.55\%                                & \textbf{97.69\%}                         & 96.48\%                              & 1.21\%                               \\
\multicolumn{1}{c|}{\textbf{}}                        & \textbf{}          & 0.25        & 187.1                                & 1.61\%                        & 14.31\%                                  & 85.69\%                                 & 97.62\%                         & \textbf{96.90\%}                              & 0.72\%                                \\
\multicolumn{1}{c|}{\textbf{}}                        & \textbf{}          & 0.5        & 160.3                                 & 1.38\%                        & 27.10\%                                  & 72.90\%                                 & 97.31\%                         & 96.80\%                              & 0.51\%                                \\
\multicolumn{1}{c|}{\textbf{}}                        & \textbf{}          & 0.75       & 178.9                                 & 1.54\%                        & 37.33\%                                 & 62.67\%                                 & 97.17\%                         & 96.66\%                              & 0.51\%                       \\
\multicolumn{1}{c|}{\textbf{}}                        & \textbf{}          & 1          & \textbf{144.6}                       & \textbf{1.24\%}               & 24.52\%                                 & 75.48\%                                 & 97.34\%                & 96.88\%                     & \textbf{0.46\%}            \\ \bottomrule
\end{tabular}%
}
\caption{Counterfactual bias insertion results on glasses bias testing with and without Targeted Data Augmentation. "p": the probability of augmentation, "switched": the number of images where the model switched its prediction when the glasses were added, "mal to fem": the percentage of images switched from female to male, "fem to mal": the percentage of images switched from male to female, "$F_1$": the overall $F_1$ score of the model on the task, "$F_1^{aug}$": the $F_1$ score of the model on augmented (biased) data. "$F_1^{diff}$": the difference between the $F_1$ score and the $F_1^{aug}$ score. This is a measure of how much the model benefits from using augmented data.}
\end{table*}

\subsection{Training details} \label{sec.training_details}
The experiments were conducted using the datasets described in \cref{sec:datasets}. For skin lesion classification, we utilized 411 ruler segmentation masks from \cite{ramella2021hair} that were made available in open repositories\footnote{\href{https://github.com/gramella/HR}{Segmentation masks of hair and rulers}}. For gender classification, we collected eyeglasses masks from various sources and published them in an open repository\footnote{Anonymous (link placeholder)}. We trained and tested three different architectures: EfficientNet-B2 \cite{tan2019efficientnet}, DenseNet121 \cite{DBLP:journals/corr/HuangLW16a}, and Vision Transformer (base version, 16 patch, 224) \cite{dosovitskiy2021an}.

All models were trained for five epochs with a learning rate of $5e-4$ for EfficientNet-B2 and DenseNet121, and $5e-5$ for ViT, respectively. The step scheduler was used to lower the learning rate every epoch by a factor of 0.9. For skin lesion classification, we used a batch size of 64, and 2 for gender classification.

Depending on the experiment scenario, we randomly inserted biases such as a ruler, a frame, and glasses with probabilities of $p=[0.0, 0.25, 0.5, 0.75, 1.0]$.

\section{Conclusions}

This paper proposes a new and effective method for mitigating biases called Targeted Data Augmentation (TDA). The experiments and results presented confirm the effectiveness of this method. In order to evaluate frame and ruler bias, 2000 skin lesion images were manually annotated. Similarly, the whole gender dataset was semi-automatically annotated using a model trained for glasses presence classification. These annotations will be publicly available for other researchers to use when evaluating bias mitigation methods in their studies. Additionally, the full code, detailed results, and used datasets with segmentation masks will be published, making it a real-world based bias mitigation benchmark\footnote{After accepting for publishing.}.

Our suspicions about bias in the datasets were objectively confirmed by training models on the datasets and testing them with the Counterfactual Bias Insertion method. The results showed that biases selected for the experiments strongly affected the models, with the frame bias having the largest $switched$ metrics, which is a commonly observed artifact strongly correlated with the malignant class. To mitigate these biases, we trained the models by randomly inserting biases during training, and we reported only the averaged results to maintain the readability of the tables and clarity of the paper.

We researched three different architectures, Densene121, EfficientNet-B2, and Vision Transformer, all of which shared similar behaviors when introduced to targeted data augmentations. All models showed a significant improvement in terms of robustness to bias after training with TDA. Notably, the Vision Transformer was strongly influenced by frame and ruler biases, and at the same time, the least affected by the glasses bias. We suspect that this is due to differences in architecture between these networks, as ViT does not use convolutional layers. The results showed that the $F_1$ score is not always the best robustness indicator. For instance, the ViT model with the highest $F_1$ values was the most affected model by the frame bias, which shows the need for careful and comprehensive model evaluation, not only by standard performance metrics.

An interesting aspect is augmentation probability $p$. In each case, $p>0$ resulted in lowering the $switched$ metric, and in most cases, $p$ within a range of $<0.25,0.75>$ gave the best results. Similarly, using TDA resulted in lowering the difference between $F_1$ and $F_1^{aug}$, which is clear evidence of immunization of the model to the bias in data. It is evident that the presented datasets are influenced by biases, and TDA allows for reducing their impact.

\section{Acknowledgements}

The  research on bias reported  in  this  publication  was supported  by  Polish  National  Science  Centre (Grant Preludium No: \textit{UMO-2019/35/N/ST6/04052}).

{\small
\bibliographystyle{ieee_fullname}
\bibliography{egbib}

\begin{thebibliography}{10}\itemsep=-1pt

\bibitem{abbas2011hair}
Qaisar Abbas, M~Emre Celebi, and Irene~Fond{\'o}n Garc{\'\i}a.
\newblock Hair removal methods: A comparative study for dermoscopy images.
\newblock {\em Biomedical Signal Processing and Control}, 6(4):395--404, 2011.

\bibitem{barata2019deep}
Catarina Barata, Jorge~S Marques, and M Emre~Celebi.
\newblock Deep attention model for the hierarchical diagnosis of skin lesions.
\newblock In {\em Proceedings of the IEEE/CVF Conference on Computer Vision and
  Pattern Recognition Workshops}, pages 0--0, 2019.

\bibitem{Bissoto_2019_CVPR_Workshops}
Alceu Bissoto, Michel Fornaciali, Eduardo Valle, and Sandra Avila.
\newblock (de)constructing bias on skin lesion datasets.
\newblock In {\em Proceedings of the IEEE/CVF Conference on Computer Vision and
  Pattern Recognition (CVPR) Workshops}, June 2019.

\bibitem{bissoto_debiasing_2020}
Alceu Bissoto, Eduardo Valle, and Sandra Avila.
\newblock Debiasing {Skin} {Lesion} {Datasets} and {Models}? {Not} {So} {Fast}.
\newblock In {\em 2020 {IEEE}/{CVF} {Conference} on {Computer} {Vision} and
  {Pattern} {Recognition} {Workshops} ({CVPRW})}, pages 3192--3201, Seattle,
  WA, USA, June 2020. IEEE.

\bibitem{chai2020human}
Chengliang Chai and Guoliang Li.
\newblock Human-in-the-loop techniques in machine learning.
\newblock {\em Data Engineering}, 37:16, 2020.

\bibitem{isic2019codella}
Noel~CF Codella, David Gutman, M~Emre Celebi, Brian Helba, Michael~A Marchetti,
  Stephen~W Dusza, Aadi Kalloo, Konstantinos Liopyris, Nabin Mishra, Harald
  Kittler, et~al.
\newblock Skin lesion analysis toward melanoma detection: A challenge at the
  2017 international symposium on biomedical imaging (isbi), hosted by the
  international skin imaging collaboration (isic).
\newblock In {\em 2018 IEEE 15th international symposium on biomedical imaging
  (ISBI 2018)}, pages 168--172. IEEE, 2018.

\bibitem{isic2019BCN20000}
Marc Combalia, Noel~CF Codella, Veronica Rotemberg, Brian Helba, Veronica
  Vilaplana, Ofer Reiter, Cristina Carrera, Alicia Barreiro, Allan~C Halpern,
  Susana Puig, et~al.
\newblock Bcn20000: Dermoscopic lesions in the wild.
\newblock {\em arXiv preprint arXiv:1908.02288}, 2019.

\bibitem{dosovitskiy2021an}
Alexey Dosovitskiy, Lucas Beyer, Alexander Kolesnikov, Dirk Weissenborn,
  Xiaohua Zhai, Thomas Unterthiner, Mostafa Dehghani, Matthias Minderer, Georg
  Heigold, Sylvain Gelly, Jakob Uszkoreit, and Neil Houlsby.
\newblock An image is worth 16x16 words: Transformers for image recognition at
  scale.
\newblock In {\em International Conference on Learning Representations}, 2021.

\bibitem{dwork2018decoupled}
Cynthia Dwork, Nicole Immorlica, Adam~Tauman Kalai, and Max Leiserson.
\newblock Decoupled classifiers for group-fair and efficient machine learning.
\newblock In {\em Conference on fairness, accountability and transparency},
  pages 119--133. PMLR, 2018.

\bibitem{he2012bias}
Jia He and Fons van~de Vijver.
\newblock Bias and equivalence in cross-cultural research.
\newblock {\em Online readings in psychology and culture}, 2(2):2307--0919,
  2012.

\bibitem{hou2018self}
Qibin Hou, PengTao Jiang, Yunchao Wei, and Ming-Ming Cheng.
\newblock Self-erasing network for integral object attention.
\newblock {\em Advances in Neural Information Processing Systems}, 31, 2018.

\bibitem{DBLP:journals/corr/HuangLW16a}
Gao Huang, Zhuang Liu, and Kilian~Q. Weinberger.
\newblock Densely connected convolutional networks.
\newblock {\em CoRR}, abs/1608.06993, 2016.

\bibitem{huang2019brain}
Qiaoying Huang, Xiao Chen, Dimitris Metaxas, and Mariappan~S Nadar.
\newblock Brain segmentation from k-space with end-to-end recurrent attention
  network.
\newblock In {\em International Conference on Medical Image Computing and
  Computer-Assisted Intervention}, pages 275--283. Springer, 2019.

\bibitem{isic2020}
{International Skin Imaging Collaboration}.
\newblock Siim-isic 2020 challenge dataset, 2020.

\bibitem{le2020adversarial}
Ronan Le~Bras, Swabha Swayamdipta, Chandra Bhagavatula, Rowan Zellers, Matthew
  Peters, Ashish Sabharwal, and Yejin Choi.
\newblock Adversarial filters of dataset biases.
\newblock In {\em International Conference on Machine Learning}, pages
  1078--1088. PMLR, 2020.

\bibitem{li2019zoom}
Hongyang Li, Yu Liu, Wanli Ouyang, and Xiaogang Wang.
\newblock Zoom out-and-in network with map attention decision for region
  proposal and object detection.
\newblock {\em International Journal of Computer Vision}, 127(3):225--238,
  2019.

\bibitem{9379034}
Miguel Luengo-Oroz, Joseph Bullock, Katherine~Hoffmann Pham, Cynthia Sin~Nga
  Lam, and Alexandra Luccioni.
\newblock From artificial intelligence bias to inequality in the time of
  covid-19.
\newblock {\em IEEE Technology and Society Magazine}, 40(1):71--79, 2021.

\bibitem{mahtani2018catalogue}
Kamal Mahtani, Elizabeth~A Spencer, Jon Brassey, and Carl Heneghan.
\newblock Catalogue of bias: observer bias.
\newblock {\em BMJ evidence-based medicine}, 23(1):23, 2018.

\bibitem{mehrabi2019survey}
Ninareh Mehrabi, Fred Morstatter, Nripsuta Saxena, Kristina Lerman, and Aram
  Galstyan.
\newblock A survey on bias and fairness in machine learning.
\newblock {\em arXiv preprint arXiv:1908.09635}, 2019.

\bibitem{mikolajczyk_towards_2021}
Agnieszka Mikołajczyk, Michał Grochowski, and Arkadiusz Kwasigroch.
\newblock Towards {Explainable} {Classifiers} {Using} the {Counterfactual}
  {Approach} - {Global} {Explanations} for {Discovering} {Bias} in {Data}.
\newblock {\em Journal of Artificial Intelligence and Soft Computing Research},
  11(1):51--67, Jan. 2021.

\bibitem{mikolajczyk_biasing_2022}
Agnieszka Mikołajczyk, Sylwia Majchrowska, and Sandra~Carrasco Limeros.
\newblock The (de)biasing effect of {GAN}-based augmentation methods on skin
  lesion images, June 2022.
\newblock Number: arXiv:2206.15182 arXiv:2206.15182 [cs, eess].

\bibitem{oliveira2016computational}
Roberta~B Oliveira, E Mercedes~Filho, Zhen Ma, Jo{\~a}o~P Papa, Aledir~S
  Pereira, and Jo{\~a}o Manuel~RS Tavares.
\newblock Computational methods for the image segmentation of pigmented skin
  lesions: a review.
\newblock {\em Computer methods and programs in biomedicine}, 131:127--141,
  2016.

\bibitem{ramella2021hair}
Giuliana Ramella.
\newblock Hair removal combining saliency, shape and color.
\newblock {\em Applied Sciences}, 11(1):447, 2021.

\bibitem{isic2020_metadata}
Veronica Rotemberg, Nicholas Kurtansky, Brigid Betz-Stablein, Liam Caffery,
  Emmanouil Chousakos, Noel Codella, Marc Combalia, Stephen Dusza, Pascale
  Guitera, David Gutman, et~al.
\newblock A patient-centric dataset of images and metadata for identifying
  melanomas using clinical context.
\newblock {\em Scientific data}, 8(1):1--8, 2021.

\bibitem{shorten2019survey}
Connor Shorten and Taghi~M Khoshgoftaar.
\newblock A survey on image data augmentation for deep learning.
\newblock {\em Journal of Big Data}, 6(1):1--48, 2019.

\bibitem{tan2019efficientnet}
Mingxing Tan and Quoc Le.
\newblock Efficientnet: Rethinking model scaling for convolutional neural
  networks.
\newblock In {\em International Conference on Machine Learning}, pages
  6105--6114. PMLR, 2019.

\bibitem{torralba2011unbiased}
Antonio Torralba and Alexei~A Efros.
\newblock Unbiased look at dataset bias.
\newblock In {\em CVPR 2011}, pages 1521--1528. IEEE, 2011.

\bibitem{isic2019ham10000}
Philipp Tschandl, Cliff Rosendahl, and Harald Kittler.
\newblock The ham10000 dataset, a large collection of multi-source
  dermatoscopic images of common pigmented skin lesions.
\newblock {\em Scientific data}, 5(1):1--9, 2018.

\bibitem{stoyanov_visualizing_2018}
Pieter Van~Molle, Miguel De~Strooper, Tim Verbelen, Bert Vankeirsbilck, Pieter
  Simoens, and Bart Dhoedt.
\newblock Visualizing {Convolutional} {Neural} {Networks} to {Improve}
  {Decision} {Support} for {Skin} {Lesion} {Classification}.
\newblock In Danail Stoyanov, Zeike Taylor, Seyed~Mostafa Kia, Ipek Oguz,
  Mauricio Reyes, Anne Martel, Lena Maier-Hein, Andre~F. Marquand, Edouard
  Duchesnay, Tommy Löfstedt, Bennett Landman, M.~Jorge Cardoso, Carlos~A.
  Silva, Sergio Pereira, and Raphael Meier, editors, {\em Understanding and
  {Interpreting} {Machine} {Learning} in {Medical} {Image} {Computing}
  {Applications}}, volume 11038, pages 115--123. Springer International
  Publishing, Cham, 2018.
\newblock Series Title: Lecture Notes in Computer Science.

\bibitem{wang2020towards}
Zeyu Wang, Klint Qinami, Ioannis~Christos Karakozis, Kyle Genova, Prem Nair,
  Kenji Hata, and Olga Russakovsky.
\newblock Towards fairness in visual recognition: Effective strategies for bias
  mitigation.
\newblock In {\em Proceedings of the IEEE/CVF conference on computer vision and
  pattern recognition}, pages 8919--8928, 2020.

\bibitem{twarz_atlas}
Karl~H. Wesker, Ralf~J. Radlanski, and Tomasz Kaczmarzyk.
\newblock {\em Twarz Atlas Anatomii Klinicznej}.
\newblock Kwintesencja, 1 edition, 2015.
\newblock PL.

\bibitem{zhang2018mitigating}
Brian~Hu Zhang, Blake Lemoine, and Margaret Mitchell.
\newblock Mitigating unwanted biases with adversarial learning.
\newblock In {\em Proceedings of the 2018 AAAI/ACM Conference on AI, Ethics,
  and Society}, pages 335--340, 2018.

\bibitem{zhao2017men}
Jieyu Zhao, Tianlu Wang, Mark Yatskar, Vicente Ordonez, and Kai-Wei Chang.
\newblock Men also like shopping: Reducing gender bias amplification using
  corpus-level constraints.
\newblock {\em arXiv preprint arXiv:1707.09457}, 2017.

\end{thebibliography}
}

\end{document}